\documentclass[onecolumn, a4paper,journal,12pt]{IEEEtran}
\usepackage{amsmath,amsfonts,amssymb}
\usepackage{graphicx}
\usepackage{setspace}
\usepackage{tocloft}
\usepackage{amsmath,graphicx}
\usepackage{amsfonts}
\usepackage{epsfig,amssymb,latexsym,graphics,float}
\usepackage{setspace,subfig}
\usepackage{citesort}
\usepackage{amsopn}
\usepackage{amssymb,times}
\usepackage{ifpdf}
\usepackage[ruled,vlined,linesnumbered]{algorithm2e}
\usepackage{setspace}
\include{IEEEtran.bib}

\begin{document}
\begin{spacing}{2.0}
\title{Robust video object tracking via Bayesian model averaging based feature fusion}
\author{Yi~Dai and Bin~Liu$^{\dag}$\\
School of Computer Science and Technology, Nanjing University of Posts and Telecommunications, Nanjing, 210023 China
\thanks{$^{\dag}$ Correspondence author. Email: bins@ieee.org. 

This work has been published online by the journal Optical Engineering since Aug. 5, 2016. So please cite it as follows 

Yi Dai, Bin Liu, ``Robust video object tracking via Bayesian model averaging-based feature fusion,” Opt.
Eng. 55(8), 083102 (2016), doi: 10.1117/1.OE.55.8.083102. }}


\maketitle
\begin{abstract}
In this article, we are concerned with tracking an object of interest in video stream. We propose an algorithm that is robust against occlusion, the presence of confusing colors, abrupt changes in the object features and changes in scale. We develop the algorithm within a Bayesian modeling framework. The state space model is used for capturing the temporal correlation in the sequence of frame images by modeling the underlying dynamics of the tracking system. The Bayesian model averaging (BMA) strategy is proposed for fusing multi-clue information in the observations. Any number of object features are allowed to be involved in the proposed framework. Every feature represents one source of information to be fused and is associated with an observation model. The state inference is performed by employing the particle filter methods.
In comparison with related approaches, the BMA based tracker is shown to have robustness, expressivity, and comprehensibility.
\end{abstract}

\begin{keywords}
Bayesian model averaging, feature fusion, particle filter, video, object tracking, LBP, texture feature
\end{keywords}

\section{Introduction}\label{sec:intro}
Object tracking in video stream \cite{yilmaz2006object} is an important task in many computer vision applications such as surveillance \cite{benfold2011stable}, augmented reality \cite{yang2015vision}, human-computer interfaces \cite{popa2015real} and medical imaging \cite{tinguely2015tracking}.
A straightforward strategy is to detect the target and determine its position frame by frame \cite{stauffer1999adaptive}. This process ignores the temporal correlation in the sequence of frame images and thus is incapable of dealing with occlusions. An alternative strategy is to use the state space to model the underlying dynamics of the tracking system. Among the numerous state space based tracking methods, the particle filter (PF), also known as Sequential Monte Carlo (SMC) method, has obtained considerable success in various kinds of
visual tracking problems. The PF methods can recursively approximate the posterior probability density function (pdf) with a set of weighted random
sampled particles which evolves conforming to the state space model. The PF method is neither limited to linear systems nor requires the noise to be Gaussian \cite{arulampalam2002tutorial,doucet2000sequential,isard1998condensation}.

Many PF-based visual trackers have been proposed in the literature. Most of them are based on a specific feature representation of the object. The commonly used features include but not limited to such as color \cite{nummiaro2003adaptive,yang2005fast}, edges \cite{vacchetti2004combining,yilmaz2006object,yang2005fast}, texture \cite{vacchetti2004combining} and motion\cite{takala2007multi}. Each feature has its own pros and cons in applications. For example, a tracker using only color feature can be robust to noise and partial occlusions, but suffers from illumination changes, or the presence of confusing colors in the scene \cite{nummiaro2003adaptive,yang2005fast}. Employing multiple features simultaneously via feature fusion methods can conceptually avoid the limitations of the single feature based methods \cite{takala2007multi,vacchetti2004combining}, while the existing fusion mechanisms are usually designed in an arbitrary manner without theoretical guarantees.

In addition, the success of such tracking methods requires an accurate object template \cite{matthews2004template}, which is usually extracted from the first frame of the video. The tracking problem can be understood as a process of finding the region which matches the template as closely as possible in the remaining frames \cite{matthews2004template}. In many algorithms, the object template keeps invariant under the assumption of that the appearance of the object remains the same throughout the entire video. This assumption may be reasonable for a certain period of time, but eventually it becomes no longer valid.
%

In this article, we introduce the concept of Bayesian model averaging (BMA) into the context of visual tracking. The BMA concept is used to fuse multi-clue information in the process of object tracking. Each clue of information is associated with one type of object feature, e.g., the color or the texture feature. A byproduct of employing BMA is shown to be an adaptive object template updating procedure, which ensures the freshness of the object template. Related multiple model based visual trackers were developed in e.g., \cite{kwon2010visual,kwon2011tracking},
while they are all based on the Markov Chain Monte Carlo (MCMC) methods, such as the Gibbs sampler, for model selection and state inference. In contrast with the existing MCMC based methods, the PF algorithm we use has significantly improved computational efficiency and less complexity in tuning parameters.

The remainder of this paper is as follows. Section 2 formulates the problem and introduces the related models. Section 3 describes the proposed BMA based feature fusion theory, along with a generic implementation of it based on the PF method. Section 4 shows the experimental results, and finally, Section 5 concludes the paper.
\section{Problem formulation and related models}
In this paper, we focus on the problem of single object tracking. We formulate the tracking problem as a Bayesian state filtering task. The aim is to estimate the conditional probability $p(X_t|Y_{0:t})$ of the target state $X_t$ at time $t$ given the sequence of observations $Y_{0:t}=(Y_0,\ldots,Y_t)$. This probability is termed the posterior distribution in the Bayesian paradigm. According to the Bayes equation, the posterior can be expressed recursively as follows
\begin{equation}\label{eqn:posterior}
p(X_t|Y_{0:t})\propto\int p(Y_t|X_t)p(X_t|X_{t-1})p(X_{t-1})|Y_{0:t-1})dX_{t-1},
\end{equation}
where the dynamic model $p(X_t|X_{t-1})$ governs the temporal evolution of the state $X_t$ given the previous state $X_{t-1}$, and the observation likelihood model $p(Y_t|X_t)$ measures the likelihood of observing $Y_t$ given the state $X_t$.

Given the dynamic and observation models (detailed in subsections \ref{sect:dyn_model} and \ref{sect:obs_model}, respectively), the task of estimating the posterior can be decomposed into a recursively processed prediction step \cite{arulampalam2002tutorial}
\begin{equation}
p(X_t|Y_{0:t-1})=\int p(X_t|X_{t-1})p(X_{t-1}|Y_{0:t-1})dX_{t-1}
\end{equation}
and update step
\begin{equation}\label{eqn:posterior2}
p(X_t|Y_{0:t})=\frac{p(Y_t|X_t)p(X_t|Y_{0:t-1})}{\int p(Y_t|X_t)p(X_t|Y_{0:t-1})dX_t}.
\end{equation}

\subsection{Dynamic model}\label{sect:dyn_model}
Given the state vector $X_t=[x_t,y_t,v_{x,t},v_{y,t},h_{x,t},h_{y,t}]$, where $[x_t,y_t]$ are the object centroid, $[v_{x,t},v_{y,t}]$ are the corresponding velocity components and $[h_{x,t},h_{y,t}]$ are width and height of the object area, the state evolution is defined as
\begin{equation}\label{dynamic model}
X_t\sim \mathcal{N}(X_t|X_{t-1},\Sigma),
\end{equation}
where the state vector $X_t$ is Gaussian distributed with mean vector $X_{t-1}$ and covariance matrix $\Sigma$. This covariance matrix is determined empirically beforehand by the model designer.
\subsection{Observation models}\label{sect:obs_model}
An observation model specifies the form of the likelihood function, which measures how likely a candidate region represents the object.
In this paper, we focus on two types of observation models by extracting two sources of information from the video stream, termed the color feature and the texture feature, respectively.
\subsubsection{Color feature based observation model}
The color feature based observation model estimates color-based similarities by using a region-based color histogram. Following Nummiaro et al. \cite{nummiaro2003adaptive}, we calculate color histograms in the RGB space using $8\times8\times8$ bins. The histograms are produced with a function $b([x,y])$, which assigns the color at location $[x,y]$ to a corresponding bin. Given the object state $X$, which defines a region covered by the object, the corresponding color distribution $p_X=\{p_X^u\}_{u=1,\ldots,U}$ over that region is calculated as follows
\begin{equation}\label{color_distribution}
p_X^u=\textbf{\mbox{C}}\sum_{j=1}^{J}\textbf{\mbox{k}}\left(\frac{\left\|X_c-[x_j,y_j]\right\|}{\sqrt{H_x^2+H_y^2}}\right)\delta(b([x_j,y_j])-u),
\end{equation}
where $\delta(\cdot)$ denotes the delta function, $U$ is the number of bins, $J$ is the number of pixels in the region of interest, $X_c$ is the object centroid corresponding to state $X$, $[H_x,H_y]$ are width and height of the region of interest, the normalization factor $\textbf{\mbox{C}}=\frac{1}{\sum_{j=1}^{J}\textbf{\mbox{k}}\left(\frac{\left\|X_c-[x_j,y_j]\right\|}{H}\right)}$
ensures that $\sum_{u=1}^{U}p_X^u=1$, and $\textbf{\mbox{k}}$ is a weighting function defined to be
\begin{equation}
\textbf{\mbox{k}}(r)= \begin{cases} 1-r^2 & 0\leq r<1\\0 &\mbox{otherwise}
\end{cases}
\end{equation}
which assigns smaller weights to the pixels that are further away from the centroid \cite{nummiaro2003adaptive}.
Denote $p_T$ to be the color distribution of an object template (see Subsection \ref{sec:template} for details about the object template), the distance between $p_X$ and $p_T$ is measured based on the Bhattacharyya distance \cite{nummiaro2003adaptive} as follows
\begin{equation}\label{dist_color}
d_{X,color}=\sqrt{1-\rho(p_X,p_T)},
\end{equation}
where $\rho(p_X,p_T)=\sum_{u=1}^U\sqrt{p_X^up_T^u}$.
The color feature based likelihood model is
\begin{equation}\label{lik_color}
p_{color}(Y|X)=\frac{1}{\sqrt{2\pi}\sigma_{color}}\exp\left(-\frac{d_{X,color}^2}{2\sigma_{color}^2}\right),
\end{equation}
where $\sigma_{color}$ has been determined empirically to be 0.1, because it is shown to be able to accommodate different scenarios in our experiments.
\subsubsection{Texture feature based observation model}
Here we focus on a local binary pattern (LBP) operator for describing texture feature. The LBP operator has been widely used in various applications such as face recognition \cite{ahonen2006face}. This operator has proven to be highly discriminative, computationally efficient and invariant to monotonic gray-level changes.

\begin{figure*}
\begin{tabular}{c}
\centerline{\includegraphics[width=6in,height=1.5in]{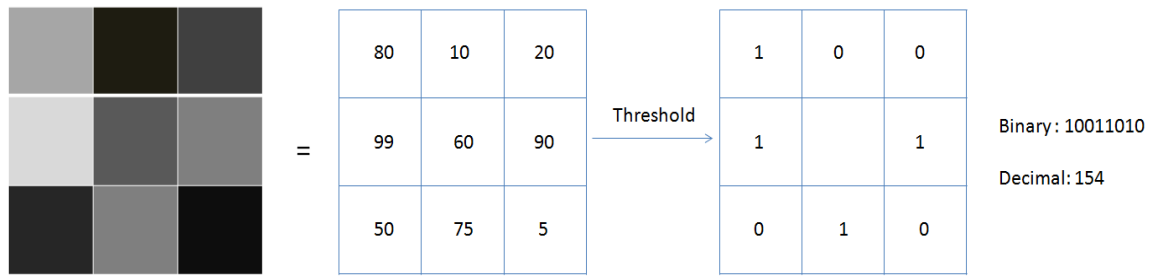}}
\end{tabular}
\caption{A basic conceptual show of the LBP operator}\label{LBP}
\end{figure*}

The LBP operator assigns a label to every pixel of an image by thresholding the $3\times3$-neighborhood of each pixel with the center pixel value. The histogram of the labels is used as a texture descriptor. A basic conceptual illustration of the LBP operator is shown in Fig.\ref{LBP}, in which the center pixel is labeled by 154.

Let $l(x,y)$ denote the label of the pixel located at $[x,y]$. A histogram of the labeled region of interest, specified by the object state $X$, can be defined as follows \cite{ye2010face}
\begin{equation}\label{texture_distribution}
H_{X,i}=\sum\limits_{x,y}I\{l(x,y)=i\}, i=0,1,\ldots,n-1
\end{equation}
where $n$ denotes the number of labels generated by the LBP operator and
\begin{equation}
I\{\mbox{A}\}= \begin{cases} 1, &\mbox{A is true}\\0, &\mbox{A is false}
\end{cases}
\end{equation}
A basic LBP operator with $n=256$ is adopted here.

Denote $H_0$ to be a LBP generated histogram corresponding to the object template (see Subsection \ref{sec:template} for details about the object template). The distance between $H_X$ and $H_0$ is measured based on the Bhattacharyya distance \cite{nummiaro2003adaptive} as follows
\begin{equation}\label{dist_texture}
d_{X,texture}=\sqrt{1-\rho(H_X,H_0)},
\end{equation}
where $\rho(H_X,H_0)=\sum_{i=1}^n\sqrt{H_{X,i}H_{0,i}}$.
The texture feature based likelihood model is just
\begin{equation}\label{lik_texture}
p_{texture}(Y|X)=\frac{1}{\sqrt{2\pi}\sigma_{texture}}\exp\left(-\frac{d_{X,texture}^2}{2\sigma_{texture}^2}\right),
\end{equation}
where $\sigma_{texture}$ has been determined empirically to be 0.1 here, as it can accommodate different scenarios in our experiments.

\section{The proposed BMA based feature fusion approach to video object tracking}
The BMA strategy is a generic solution to deal with model uncertainty problems in a Bayesian statistical paradigm \cite{hoeting1999bayesian,raftery1997bayesian,wintle2003use}.
Here we propose a BMA based feature fusion theory, along with a generic implementation of it based on the PF method, in the context of visual tracking.
\subsection{BMA based feature fusion}
We focus on the situation where several candidate features (such as the color and texture features presented in Section \ref{sect:obs_model}) are available for use, but there is uncertainty on the best feature to use at each time step. Associating each feature with a plausible model, the BMA strategy is used to balance usages of the candidate models in a theoretically sound manner.

Let $\mathcal{H}_t=m$ denote the event that the $m$th model,
$\mathcal{M}_m$, is the best for use at time step $t$.
Based on BMA \cite{hoeting1999bayesian,raftery1997bayesian,wintle2003use}, the posterior distribution, as shown in Eqns. (\ref{eqn:posterior}) and (\ref{eqn:posterior2}), can be calculated as follows
\begin{eqnarray}\label{MM_distribution3}
p(X_t|Y_{0:t})&=&\sum\limits_{m=1}^Mp(X_t|\mathcal{H}_t=m,Y_{0:t}) p(\mathcal{H}_t=m|Y_{0:t})\nonumber\\
&=&\sum\limits_{m=1}^Mp_m(X_t|Y_{0:t})\pi_{t|t,m}
\end{eqnarray}
where $p_m(X_t|Y_{0:t})\triangleq p(X_t|\mathcal{H}_t=m,Y_{0:t})$, $\pi_{t|t,m}\triangleq p(\mathcal{H}_{t}=m|Y_{0:t})$ and $M$ is the number of candidate models. Here we only consider two candidate models, namely the color and texture feature based observation models as pressented in subsection \ref{sect:obs_model}; so we have $M=2$. Note that all the calculations presented in what follows are valid for any value of $M$, $M\in\mathbb{R}^{+}$.

We use the PF method \cite{arulampalam2002tutorial,doucet2000sequential} to calculate Eqn.(\ref{MM_distribution3}). Assume that at time $t-1$, we have at hand $\pi_{t-1|t-1,m}$ and a weighted sample set, $\{X_{t-1}^i,\omega_{m,t-1}^i\}, i=1,2,\ldots, N$, which can build up a discrete probability distribution that approximates $p_m(X_{t-1}|Y_{0:t-1})$ as follows
\begin{equation}\label{eqn:particle_approx_t-1}
p_m(X_{t-1}|Y_{0:t-1})\simeq\sum_{i=1}^N\omega_{m,t-1}^i\delta(X_{t-1}-X_{t-1}^i),
\end{equation}
where $\omega_{m,t-1}^i>0, i=1,2,\ldots,N$ and $\sum_{i=1}^N\omega_{m,t-1}^i=1$, for $\forall m$.
Then the posterior at time $t-1$ can be approximated as follows
\begin{eqnarray}\label{MM_distribution2}
p(X_{t-1}|Y_{0:t-1})&=&\sum\limits_{m=1}^Mp_m(X_{t-1}|Y_{0:t-1})\pi_{t-1|t-1,m}\nonumber\\
&\simeq & \sum\limits_{m=1}^M \pi_{t-1|t-1,m} \sum_{i=1}^N \omega_{m,t-1}^i
\delta(X_{t-1}-X_{t-1}^i).
\end{eqnarray}

Comparing Eqn.(\ref{MM_distribution2}) with Eqn.(\ref{MM_distribution3}), we can observe that, upon the arrival of $Y_t$, the task of calculating $p(X_t|Y_{0:t})$ can be decomposed into the following two sub-tasks,
\begin{itemize}
 \item sub-task I:  given $\{X_{t-1}^i,\omega_{m,t-1}^i\}, i=1,2,\ldots, N$, how to generate another weighted sample set $\{X_{t}^i,\omega_{m,t}^i\}, i=1,2,\ldots, N$, which can provide a Monte Carlo approximation to $p_m(X_t|Y_{0:t})$ as follows,
     \begin{equation}\label{MM_distribution4}
     p_m(X_t|Y_{0:t})\simeq\sum_{i=1}^N\omega_{m,t}^i\delta(X_{t}-X_{t}^i), m=1,\ldots,M.
     \end{equation}
\item sub-task II: given $\pi_{t-1|t-1,m}$, how to derive $\pi_{t|t,m}$ out, for $\forall m$.
\end{itemize}
In what follows, we present solutions to these sub-tasks.
\subsubsection{PF based solution to sub-task I}\label{sec:PF_task1}
For any, say the $m$th, candidate model, $\mathcal{M}_m$, here the concern is, given $\{X_{t-1}^i,\omega_{m,t-1}^i\}, i=1,2,\ldots, N$ that satisfies Eqn.(\ref{eqn:particle_approx_t-1}), how to generate another weighted sample set $\{X_{t}^i,\omega_{m,t}^i\}, i=1,2,\ldots, N$, which should satisfy Eqn.(\ref{MM_distribution4}).

Within the PF algorithm framework, the new state samples $X_{t}^i$ are first drawn from a proposal distribution $q(X_t|X_{1:t-1},Y_{1:t})$ and then weighted according to the importance sampling strategy\cite{arulampalam2002tutorial,doucet2000sequential}. Here the state transitional prior, as defined in Eqn.(\ref{dynamic model}), is selected as the proposal, i.e., $q(X_t|X_{1:t-1},Y_{0:t})=p(X_t|X_{t-1})$. This type of proposal has been widely adopted in PF methods such as the condensation method and the bootstrap filter \cite{isard1998condensation,smith2013sequential}. Based on Eqn.(\ref{dynamic model}), we generate $X_{t}^i, i=1,\dots,N$ as follows
\begin{equation}\label{eqn:state_trans}
X_t^i \sim \mathcal{N}(X_t|X_{t-1}^i,\Sigma), i=1,\ldots,N.
\end{equation}
The corresponding importance weights are calculated as follows\cite{arulampalam2002tutorial}
\begin{eqnarray}\label{eqn:unnormalized_weight}
\hat{\omega}_{m,t}^i&=&\omega_{m,t-1}^i\frac{p_m(Y_t|X_t^i)p(X_t^i|X_{t-1}^i)}{q(X_t^i|X_{1:t-1}^i,Y_{0:t})}\nonumber\\
&=&\omega_{m,t-1}^ip_m(Y_t|X_t^i), i=1,\ldots,N.
\end{eqnarray}
\begin{equation}\label{eqn:normalized_weight}
\omega_{m,t}^i=\frac{\hat{\omega}_{m,t}^i}{\sum_{j=1}^N\hat{\omega}_{m,t}^j}, i=1,\ldots,N.
\end{equation}
As the operators for generating and weighting samples belong to the routine PF framework, the results have theoretical guarantees as proved in the literature \cite{crisan2002survey,hu2008basic}.
\subsubsection{Solution to sub-task II}
Here we focus on the following task, namely, given $\pi_{t-1|t-1,m}$, how to derive $\pi_{t|t,m}$ out, for $\forall m$.
First, we consider the prediction of the model indicator, namely given $\mathcal{H}_{t-1}$, how to predict $\mathcal{H}_t$.
We specify the model transition process in term of forgetting \cite{liu2011instantaneous}.
Denote $\alpha$ as a forgetting factor satisfying $0<\alpha<1$. Given $\pi_{t-1|t-1,m}$, $\pi_{t|t-1,m}\triangleq p(\mathcal{H}_{t}=m|Y_{0:t-1})$ is calculated as follows
\begin{equation}\label{model_pred_forget}
\pi_{t|t-1,m}=\frac{\pi_{t-1|t-1,m}^{\alpha}}{\sum_{l=1}^M\pi_{t-1|t-1,l}^{\alpha}}.
\end{equation}
Then, employing the Bayes' rule we have
\begin{equation}\label{posterior_model_indicator}
\pi_{t|t,m}=\frac{\pi_{t|t-1,m}p_m(Y_t|Y_{0:t-1})}{\sum\limits_{l=1}^M\pi_{t|t-1,l}p_l(Y_t|Y_{0:t-1})},
\end{equation}
where $p_m(Y_t|Y_{0:t-1})$ is the marginal likelihood of
$\mathcal{M}_m$ at time $t$, which is defined to be
\begin{equation}\label{marginal_lik}
p_m(Y_t|Y_{0:t-1})=\int p_m(Y_t|X_t)p_m(X_t|Y_{0:t-1})dX_t.
\end{equation}
The element $p_m(X_t|Y_{0:t-1})$ in Eqn.(\ref{marginal_lik}) can be estimated as follows,
\begin{equation}
p_m(X_t|Y_{0:t-1})\approx\sum_{i=1}^N\omega_{m,t-1}^i\delta(X_t-X_t^i),
\end{equation}
where the weighted sample set $\{\omega_{m,t-1}^i,X_t^i\}, i=1,2,\ldots, N$ is a byproduct of the PF solution to sub-task I mentioned above.
Therefore we can estimated the integral in Eqn.(\ref{marginal_lik}) as follows
\begin{equation}\label{eqn:particle_marignal_lik}
p_m(Y_t|Y_{0:t-1})\simeq\sum\limits_{i=1}^N \omega_{m,t-1}^ip_m(Y_t|X_t^i).
\end{equation}
\subsection{Updating object template}\label{sec:template}
Both the definitions of the color and texture feature based observation models require a pre-determined object template, as shown in Eqns.(\ref{dist_color}) and (\ref{dist_texture}), respectively. Based on the assumption of that the appearance of the object remains the same throughout the entire video, an invariant object template is used in many methods. This assumption may be reasonable for a certain period of time, but eventually the template will become no longer an accurate model of the appearance of the object.

Here we show that, as a byproduct of employing BMA for object tracking, an adaptive template updating mechanism can be easily realized to ensure that the current template accurately represents the new image of the object.

In our methods, the initial object template is produced by an object detector \cite{stauffer1999adaptive}.
This detector employs an adaptive Gaussian mixture to model the time-evolving scene in the video stream .
During the follow-up tracking process, the color or texture distribution of a predicted object region is enforced to be compared with that of the object template to determine the likelihood of the new observation via Eqn.(\ref{lik_color}) or (\ref{lik_texture}).
If an abrupt change in one feature space of the object happens, it will result in a sudden slump in the corresponding likelihood.
Then the marginal likelihood and posterior probability of that feature based observation model, in terms of Eqn.(\ref{marginal_lik}) or Eqn.(\ref{posterior_model_indicator}), will be reduced correspondingly.
Therefore, the state estimate produced by using that feature will be assigned an extremely small probability weight in generating the final state estimate by Eqn.(\ref{MM_distribution3}). Therefore, the BMA based method can be robust to the failure of a single feature based model in yielding accurate estimation of the object state.

The template updating procedure can be simple as follows. If a slump in the posterior probability of a feature is observed, we consider it as an indication of that we need to update the object template. We extract a new image of the object based on the output of the BMA based tracker, and then construct a new object template model. If it is the color (or texture) feature based observation model that fails, we construct the new object template model by calculating the feature distribution by Eqn.(\ref{color_distribution})(or Eqn.(\ref{texture_distribution})).
\subsection{Implementation of the proposed algorithm}
Here a particle based implementation of the proposed method is summarized as follows in \textbf{Algorithm 1}.
In this implementation two models are considered and thus $m\in\{1,2\}$, where the figures 1 and 2 correspond to the color and texture feature models, respectively.
\begin{algorithm}[!htb]
Input: the '\textit{old}' sample set $\{\omega_{1,t-1}^i, \omega_{2,t-1}^i, X_{t-1}^i\}, i=1,2,\ldots, N$, and the '\textit{old}' posterior probabilities of the candidate models $\pi_{t-1|t-1,1}$ and $\pi_{t-1|t-1,2}$, at time step $t-1$; the color distribution $p_T$ and the LBP generated histogram $H_0$ of the object template\;
\For{$i=1,\ldots, N$ }{
    Sample $X_{t}^i$ using Eqn.(\ref{eqn:state_trans})\;
    Calculate the importance weights $\hat{\omega}_{m,t}^i, m=1,2$ using Eqn.(\ref{eqn:unnormalized_weight}), in which $p_1(Y_t|X_t^i)$ and $p_2(Y_t|X_t^i)$ are replaced with $p_{color}(Y_t|X_t^i)$ (defined by Eqn.(\ref{lik_color})) and $p_{texture}(Y_t|X_t^i)$ (defined by Eqn.(\ref{lik_texture})), respectively\;
}
Normalize the importance weights using Eqn. (\ref{eqn:normalized_weight}), and get ${\omega}_{m,t}^i, i=1,2,\ldots, N, m=1,2$ \;
Calculate $\pi_{t|t-1,m}, m=1,2,$ using Eqn. (\ref{model_pred_forget})\;
Calculate $\pi_{t|t,m}, m=1,2,$ using Eqns. (\ref{posterior_model_indicator}-\ref{eqn:particle_marignal_lik}), in which $p_1(Y_t|X_t^i)$ and $p_2(Y_t|X_t^i)$ are replaced with $p_{color}(Y_t|X_t^i)$ (defined by Eqn.(\ref{lik_color})) and $p_{texture}(Y_t|X_t^i)$ (defined by Eqn.(\ref{lik_texture})), respectively\;
Estimate, if desired, moments of the tracked position at time step $t$ as
$\xi(f(X_t))=\sum_{m=1}^M\pi_{t|t,m}\sum_{i=1}^N{\omega}_{m,t}^if(X_t^i)$, obtaining, for instance, a mean position using $f(X)=X$\;
If the condition of updating the object template satisfies, update $p_T$ or $H_0$ correspondingly, based on the current estimate of the object state, see details in Subsection \ref{sec:template}\;
Output: $\xi(f(X_t))$, $\{\omega_{1,t}^i, \omega_{2,t}^i, X_{t}^i\}, i=1,2,\ldots, N$, $\pi_{t|t,1}$, $\pi_{t|t,2}$, $p_T$ and $H_0$.
\caption{\label{algo:BMA-PF}One iteration of the proposed BMA-PF algorithm}
\end{algorithm}
%
\section{Experimental Results}
We applied the proposed method to analyze real video stream data.
The purpose is to demonstrate that the proposed BMA based feature fusion method really works.

Several competitor algorithms, including ALG I (PF using adaptive color feature \cite{nummiaro2003adaptive}), ALG II (PF using LBP texture feature \cite{ye2010face}), ALG III (PF using both the texture and color features, which are equally weighted \cite{ying2010particle}) and ALG IV (adaptive GM detector \cite{stauffer1999adaptive}), are involved for performance comparison. The particle size $N$ in each algorithm is set equally to be 200.
\subsection{Case I}\label{sec:case1}
The first video stream under investigation is taken by a camera placed at a fixed location in a dark tunnel. The window size of the video keeps fixed as the object appears nearer or further in the frames. The task is to track a moving car passing through the tunnel.
The color of the car is similar with the road in the video. In the last frames, the taillights of the car light up leading to a change in the color feature distribution of the object.

\begin{figure*}
\begin{tabular}{c}
\centerline{\includegraphics[width=5.5in,height=3.6in]{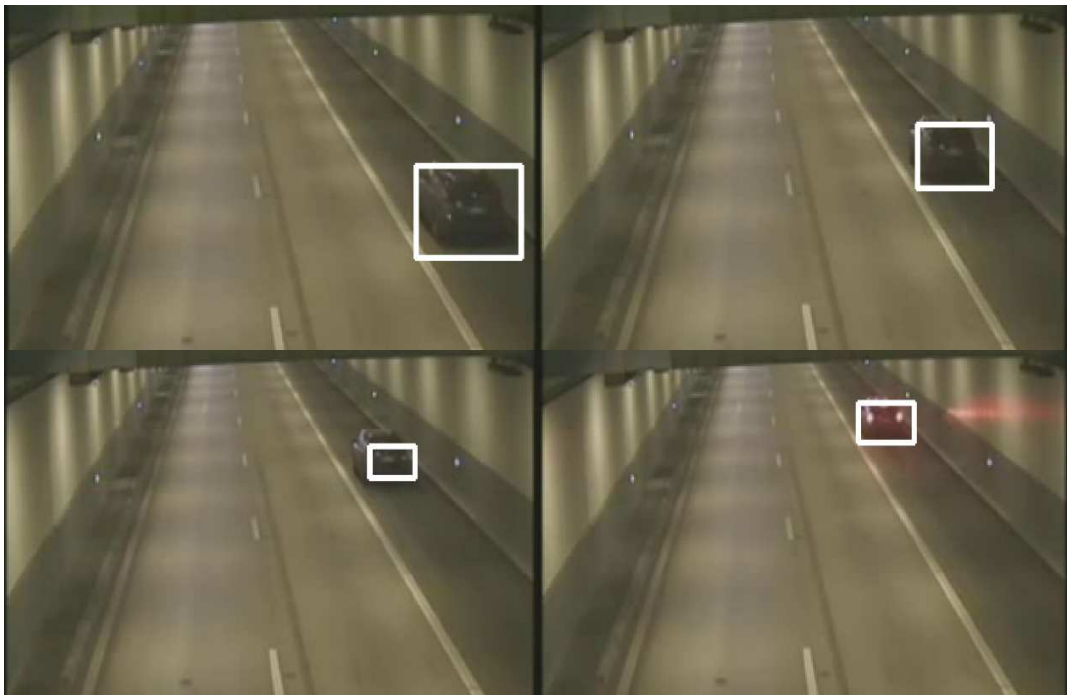}}
\end{tabular}
\caption{Tracking results of the proposed BMA algorithm for case I. In the upper left sub-figure, this car has just entered the surveillance region. The upper right and lower left sub-figures show the middle process of the surveillance. In the lower right sub-figure, the taillights of the car has just lighted up. The white box indicates the object contour outputted by the algorithm.} \label{case1_track}
\end{figure*}

\begin{figure*}
\begin{tabular}{c}
\centerline{\includegraphics[width=5.5in,height=3.6in]{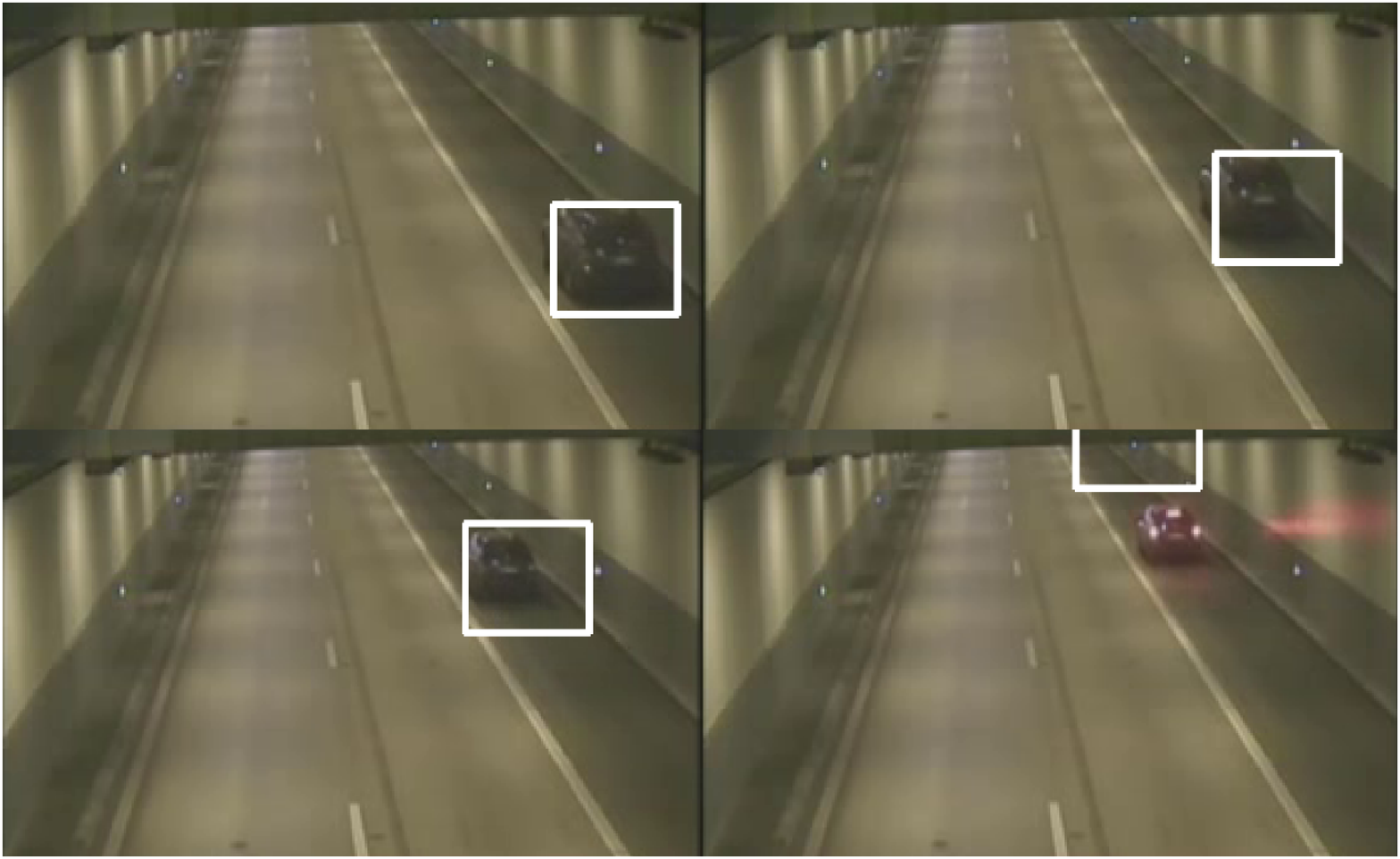}}
\end{tabular}
\caption{Tracking results of ALG I, namely the PF tracker using adaptive color feature \cite{nummiaro2003adaptive}, for case I.} \label{case1_color_track}
\end{figure*}

\begin{figure*}
\begin{tabular}{c}
\centerline{\includegraphics[width=5.5in,height=3.6in]{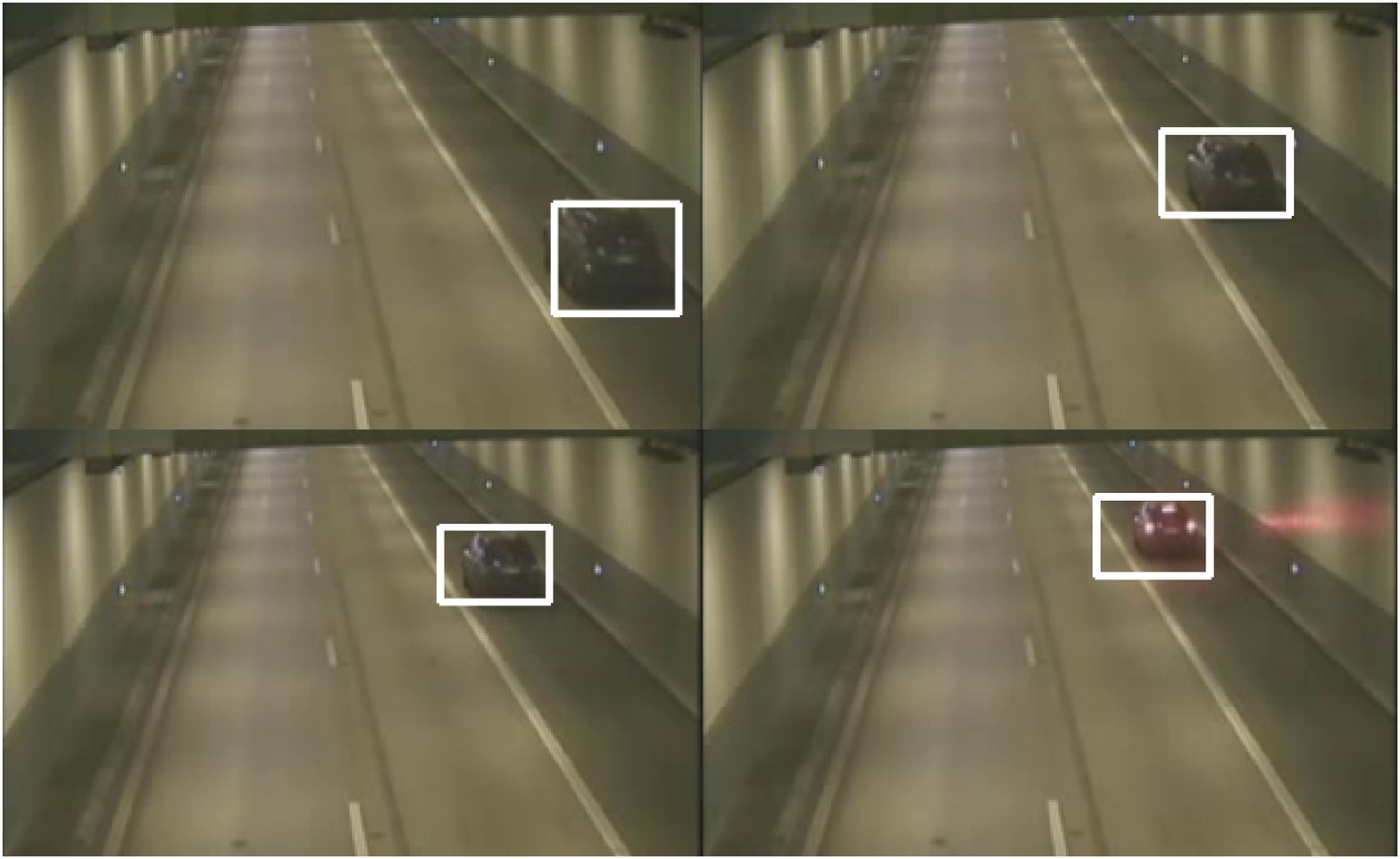}}
\end{tabular}
\caption{Tracking results of ALG II, namely the PF tracker using LBP modeled texture feature \cite{ye2010face}, for case I.} \label{case1_texture_track}
\end{figure*}

\begin{figure*}
\begin{tabular}{c}
\centerline{\includegraphics[width=5.5in,height=3.6in]{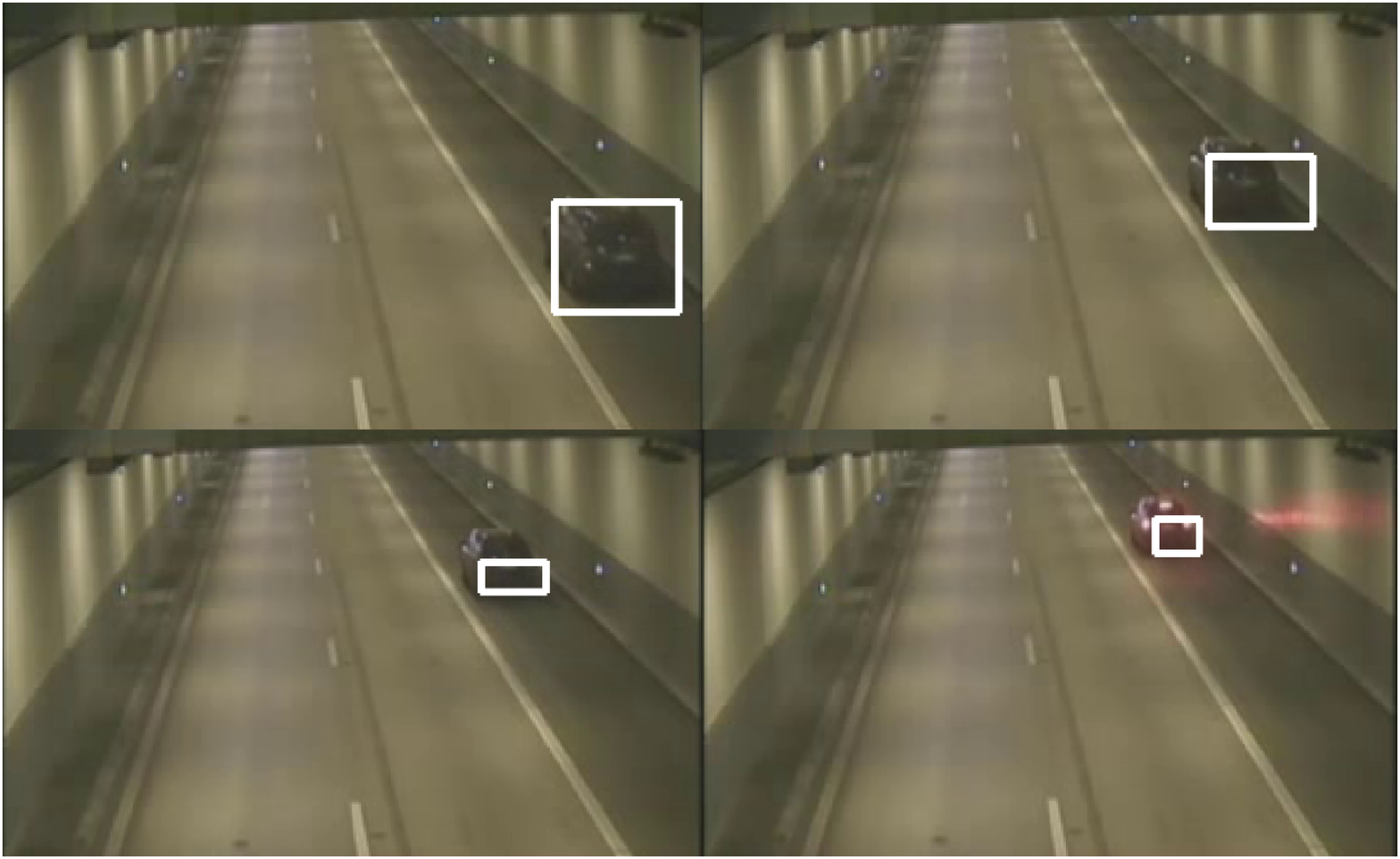}}
\end{tabular}
\caption{Tracking results of ALG III, namely the PF tracker using equally weighted texture and color features \cite{ying2010particle}, for case I.} \label{case1_fw_track}
\end{figure*}

\begin{figure*}
\begin{tabular}{c}
\centerline{\includegraphics[width=5.5in,height=3.6in]{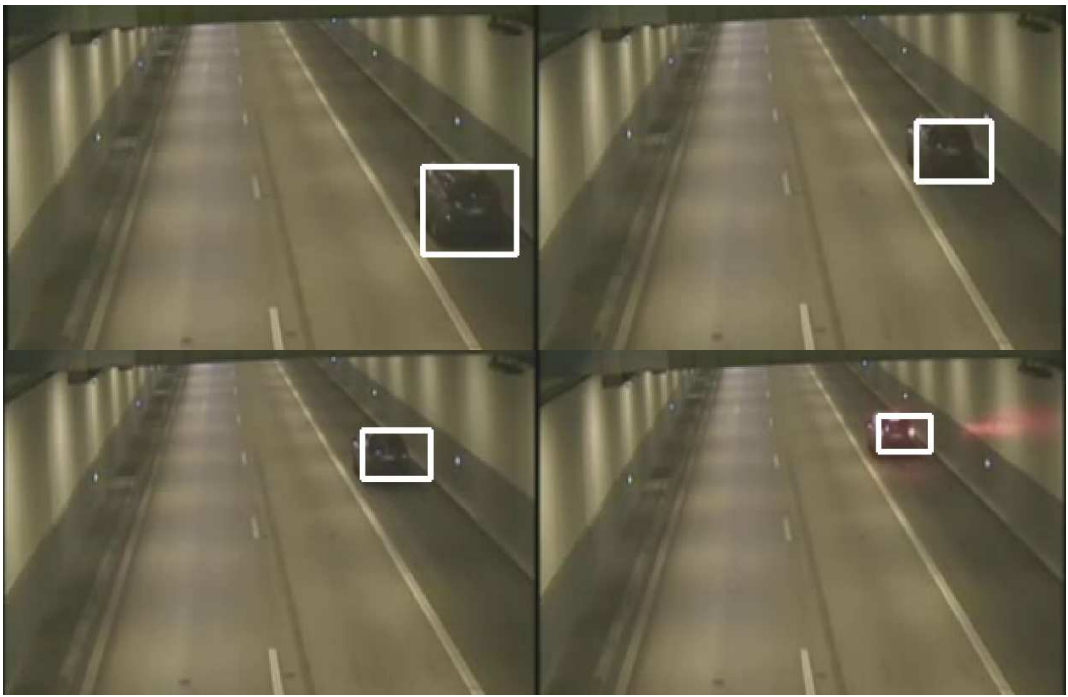}}
\end{tabular}
\caption{Tracking results of ALG IV, namely the adaptive GM detector \cite{stauffer1999adaptive}, for case I.} \label{case1_mog_track}
\end{figure*}

The result of an example run of the proposed algorithm is presented in Fig.\ref{case1_track}. As is shown, the tracking result was not influenced by the presence of confusing colors, abrupt changes in the object color feature distribution and changes in scale.
The change in the object template has been indicated by the change in the size of the white box, which means the object contour outputted by the algorithm. For comparison, the tracking results corresponding to the competitor algorithms are shown in Figs. \ref{case1_color_track}-\ref{case1_mog_track}, respectively. We see that, among the competitor algorithms, ALG I and II did not adapt well to the change in scale. ALG I failed to track after that the taillights of the car light up.
ALG III and IV provide satisfactory tracking result.

\begin{figure*}
\begin{tabular}{c}
\centerline{\includegraphics[width=5.5in,height=3.6in]{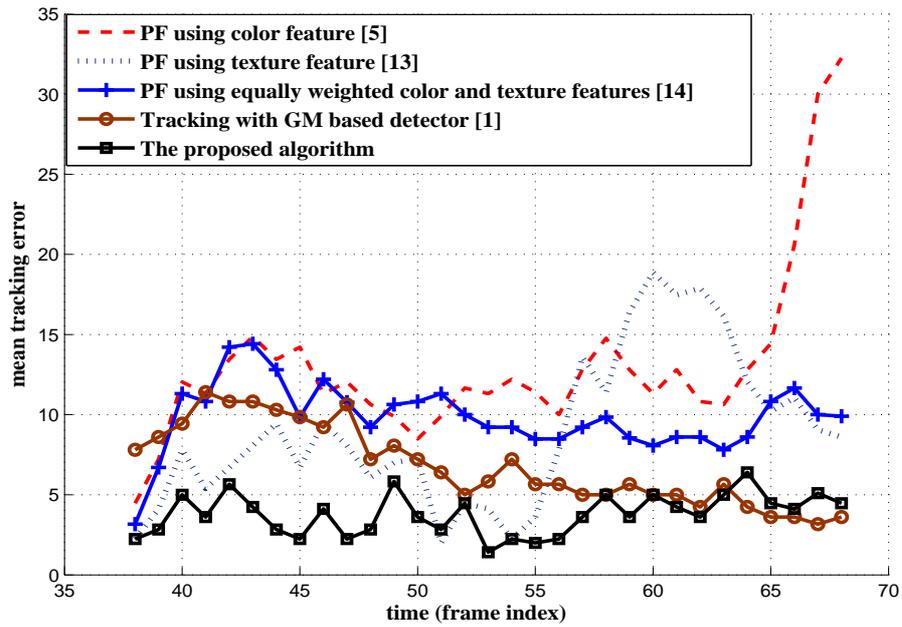}}
\end{tabular}
\caption{Mean tracking error in test case I. The object appears in the 38th frame. The size of a whole image in one frame is 320$\times$240.} \label{error_caseI}
\end{figure*}

A numerical performance comparison based on 100 times independent runs of each involved algorithm is conducted and the result is shown in Fig.\ref{error_caseI}.
We can see that the proposed algorithm performs best, while the performance of ALG I \cite{nummiaro2003adaptive} gets deteriorated remarkably since the car's taillights light up.
The computation time required to run each algorithm is presented in Table \ref{table_comp}.

\begin{figure*}
\begin{tabular}{c}
\centerline{\includegraphics[width=5.5in,height=3.0in]{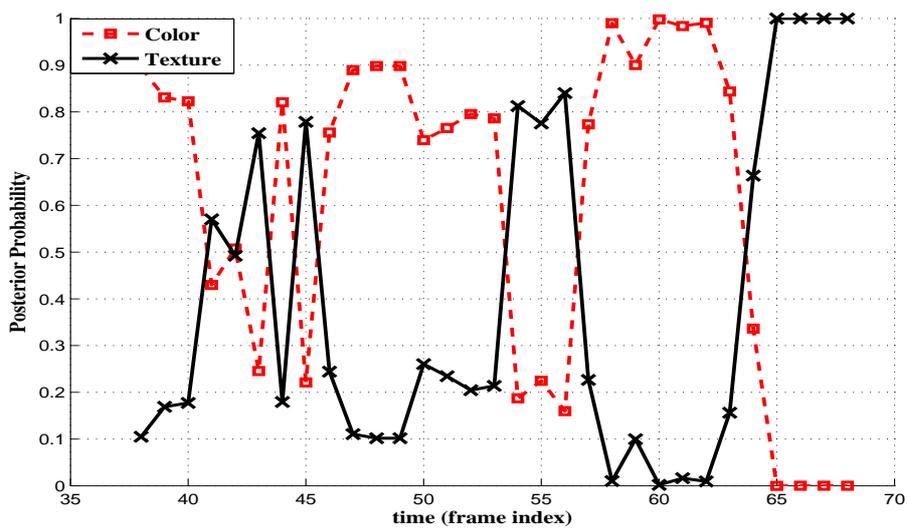}}
\end{tabular}
\caption{Posterior probabilities of the involved feature models in test case I.} \label{prob_feature_caseI}
\end{figure*}

The feature fusion result can be revealed by the changes in the posterior probabilities of the color and texture based observation models.  Fig.\ref{prob_feature_caseI} shows the real-time output of the posterior probabilities of those two models.
It is shown that the impact of the red light on the color and texture clues is significant, and that the feature fusion effect of our algorithm is truly taking effect in dynamically adjusting the usages of the color and texture features in the tracking process.
\subsection{Case II}\label{sec:case2}
\begin{figure*}
\begin{tabular}{c}
\centerline{\includegraphics[width=5.5in,height=3.6in]{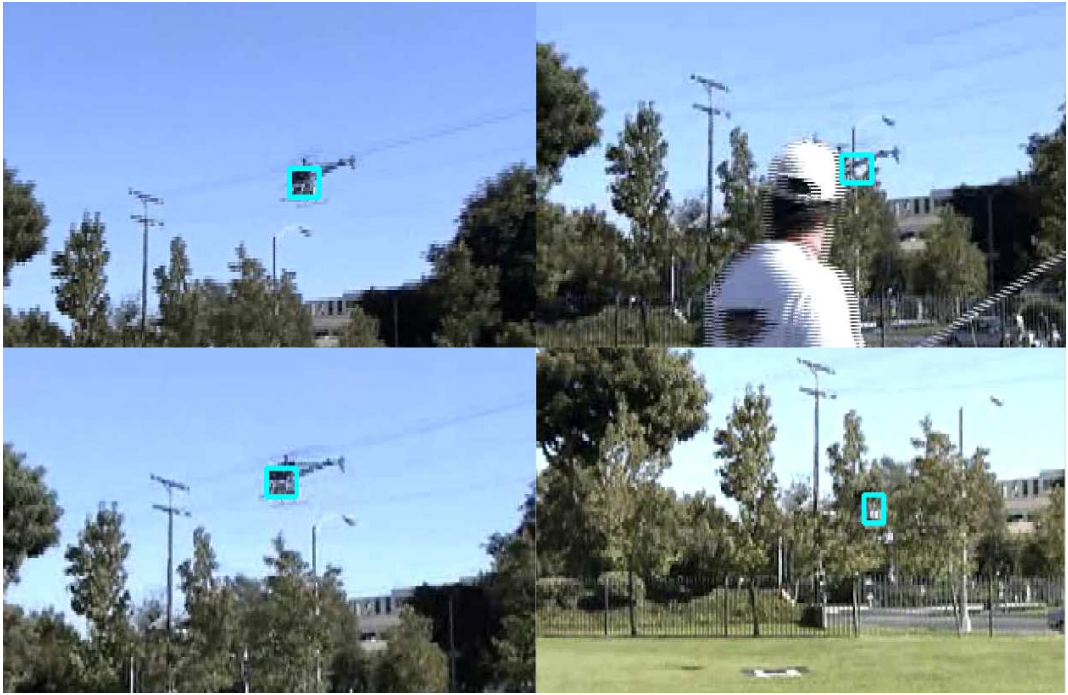}}
\end{tabular}
\caption{Tracking results of the proposed BMA algorithm for case II} \label{case2_track}
\end{figure*}

To further demonstrate the performance of the proposed method, we applied it to analyze another video stream.
The related camera was placed at a fixed location, so the window size of the video keeps fixed as the object appears nearer or further in the frames.
The object to be tracked is a green helicopter, which is controlled to fly up and down.
Four typical frames along with the tracking result provided by our algorithm are listed in Fig.\ref{case2_track}. In the upper left sub-figure, we see that the helicopter is flying above the trees. Then it flies downwards and then gets partially occluded by the trees in the upper right sub-figure. The helicopter rises up again and then appears in the lower left sub-figure. Then it falls downwards again, with its body mixed up with the trees behind it in the video.

\begin{figure*}
\begin{tabular}{c}
\centerline{\includegraphics[width=5.5in,height=3.6in]{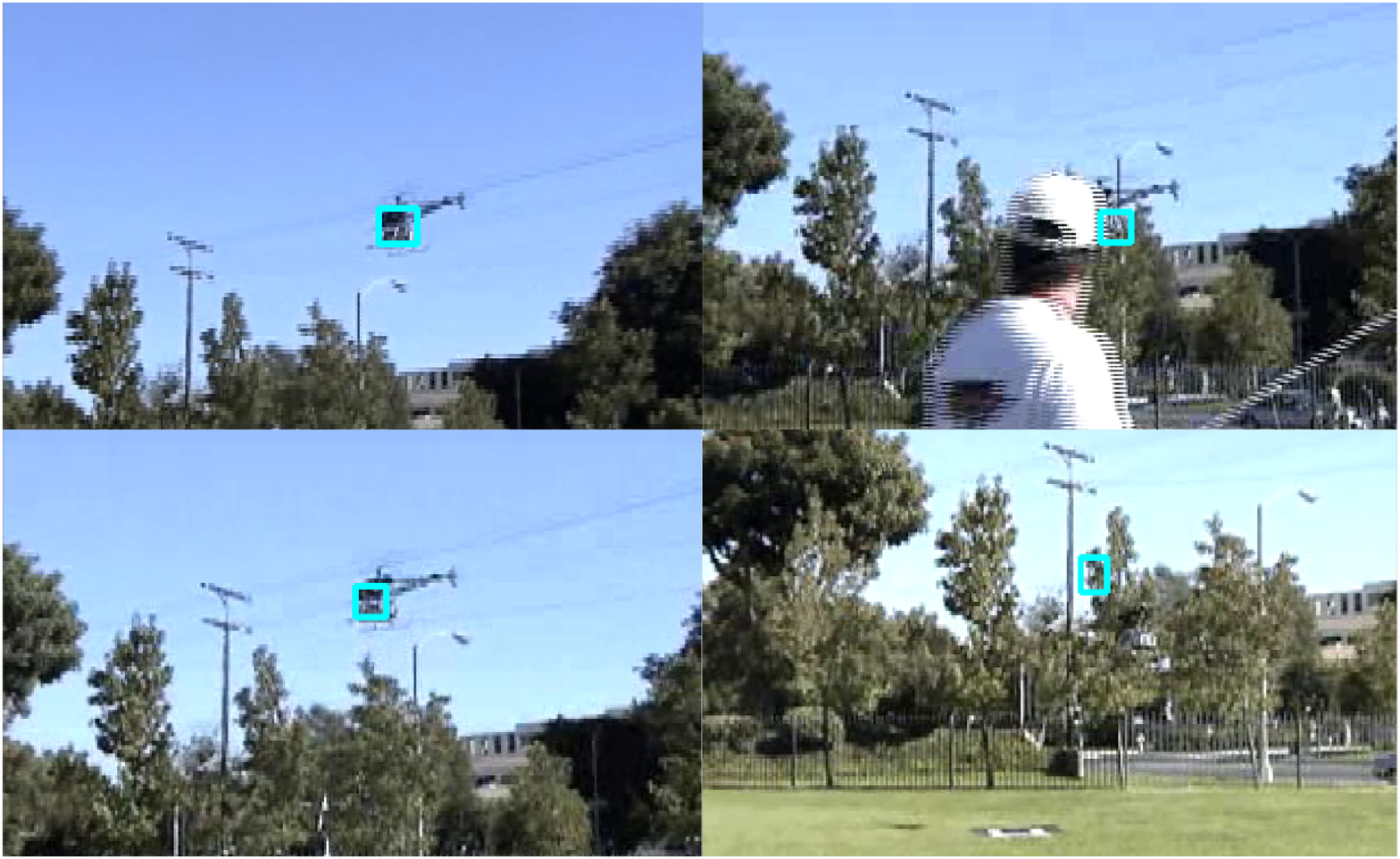}}
\end{tabular}
\caption{Tracking results of ALG I, namely PF tracker using adaptive color feature \cite{nummiaro2003adaptive}, for case II} \label{case2_color_track}
\end{figure*}

\begin{figure*}
\begin{tabular}{c}
\centerline{\includegraphics[width=5.5in,height=3.6in]{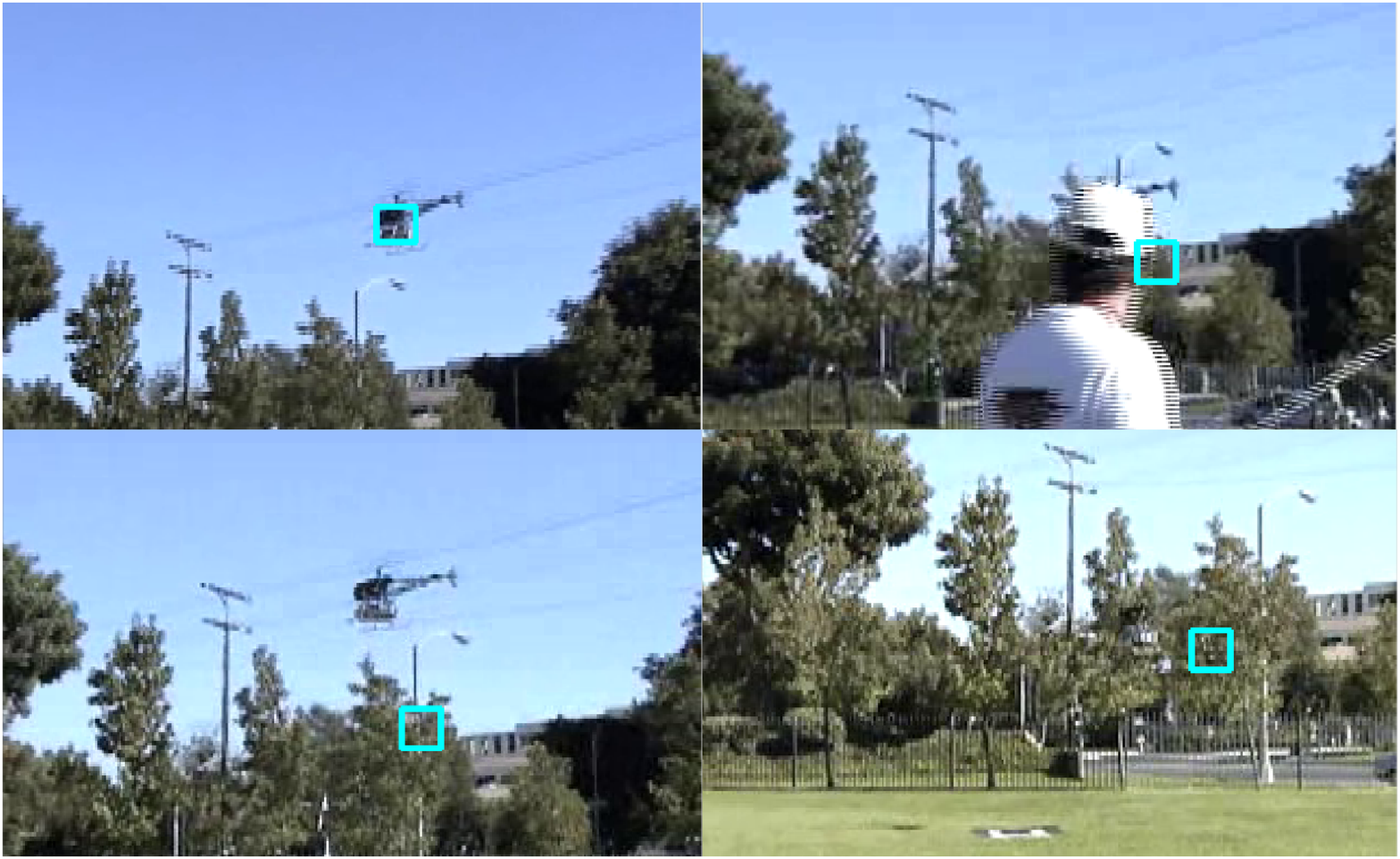}}
\end{tabular}
\caption{Tracking results of ALG II, namely the PF tracker using the LBP modeled texture feature \cite{ye2010face}, for case II} \label{case2_texture_track}
\end{figure*}

\begin{figure*}
\begin{tabular}{c}
\centerline{\includegraphics[width=5.5in,height=3.6in]{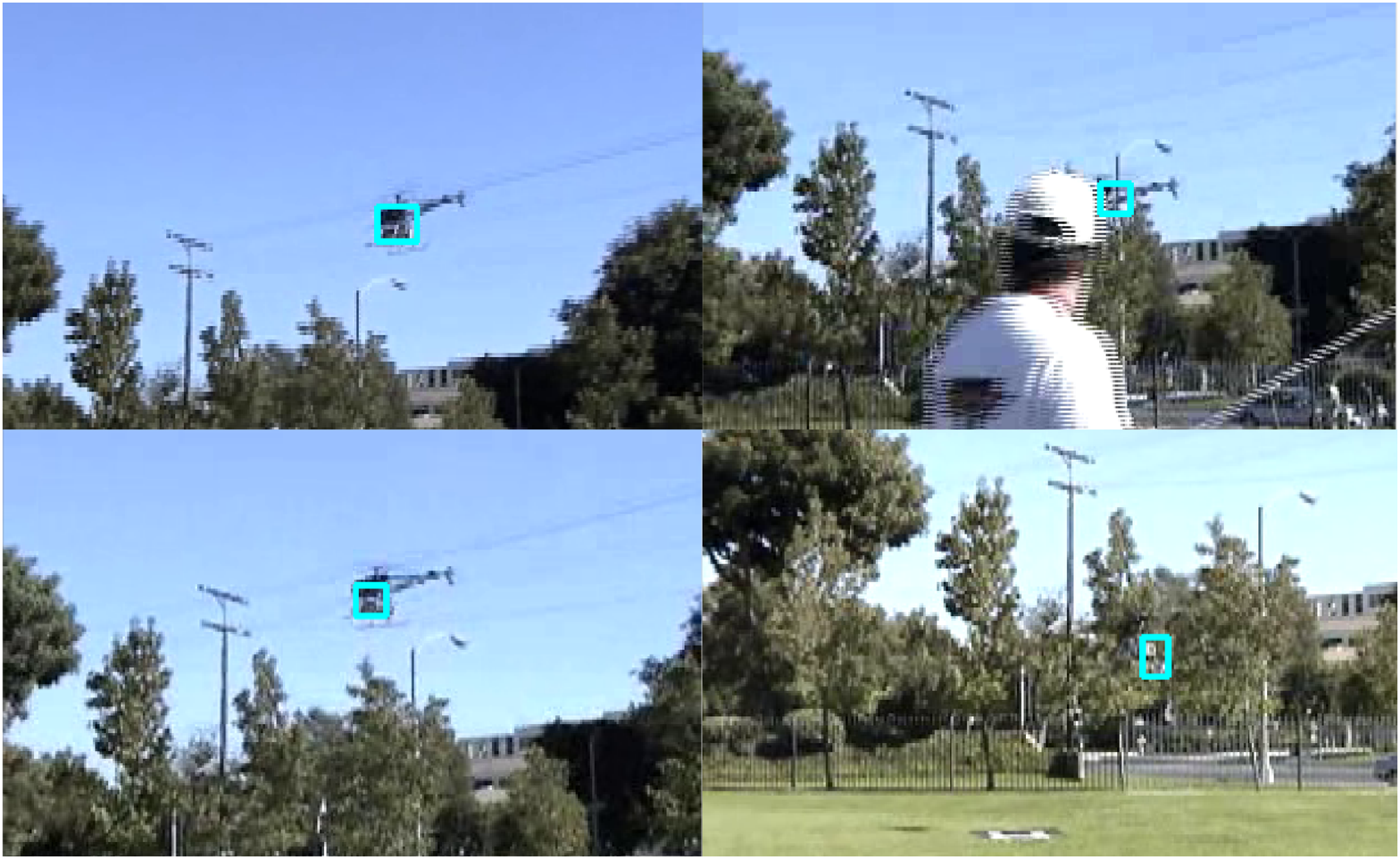}}
\end{tabular}
\caption{Tracking results of ALG III, namely the PF tracker using equally weighted texture and color features \cite{ying2010particle}, for case II} \label{case2_fw_track}
\end{figure*}

In this case, the proposed algorithm worked very well in tracking the object accurately from beginning to end. See Fig.\ref{case2_track} for 4 typical frames in the tracking period. As shown in Fig. \ref{case2_color_track}, ALG I \cite{nummiaro2003adaptive} failed when the body of the object is totally mixed up with the trees in the lower right sub-figure, because the algorithm erroneously identified the crown of a tree as the object. For ALG II \cite{ye2010face}, an early tracking failure occured when the object approaches the trees in the first time, see the upper right sub-figure of Fig. \ref{case2_texture_track}. In Fig.\ref{case2_fw_track}, we see that ALG III \cite{ying2010particle} performed satisfactorily for this case.
\subsection{Case III: PETS 2015 dataset}\label{sec:case3}
To further properly evaluate the proposed method, we applied it to analyze an open source dataset released in the 2015 International Workshop on Performance Evaluation of Tracking and Surveillance (PETS 2015) \cite{li2015pets}. We selected one dataset that conforms to the application scenario of the presented methods. This dataset is called P5, whose acronym stands for the EU project 'Privacy Preserving Perimeter Protection Project'.

The object to be tracked is a vehicle driving across the scene. This vehicle appears from the bottom right corner of the scene. It first turns right along with a riverside path. Then the vehicle moves ahead to the direction far away from the camera. The shape, contour and size of the object continue to change over time in this video. As object's relative position changes, the object's color (especially the color of its roof) also changes gradually over time. The background is complex, containing a wide blue river, green trees, yellow grasses and some other entities like boats, rocks, and so on.

\begin{figure*}
\begin{tabular}{c}
\centerline{\includegraphics[width=5.5in,height=3.6in]{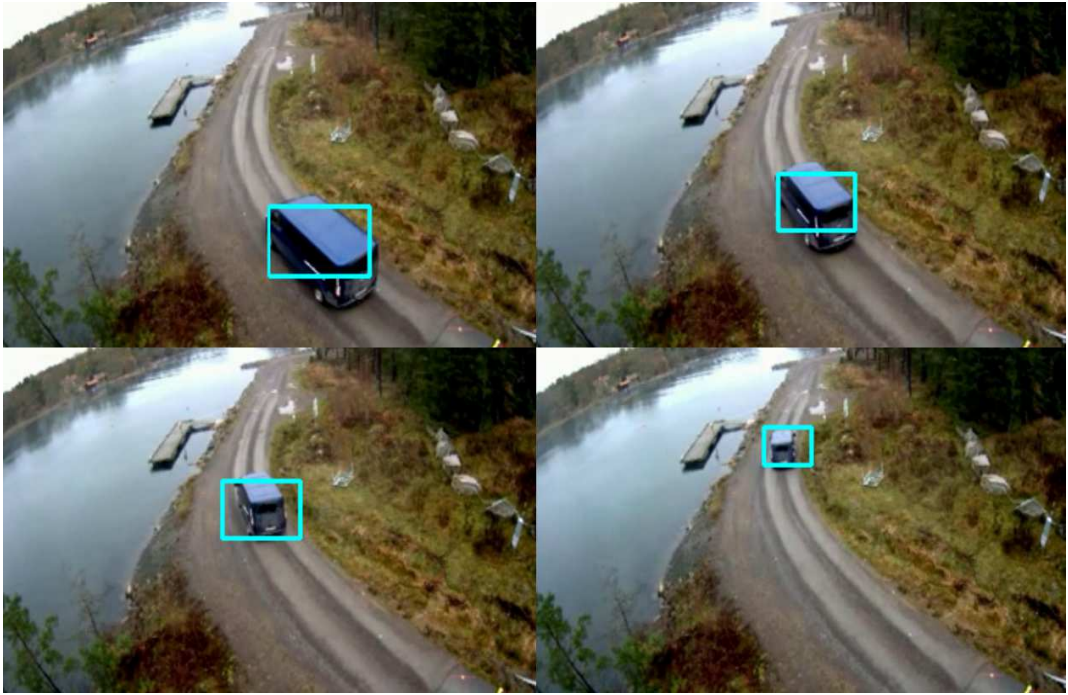}}
\end{tabular}
\caption{Tracking results of the proposed algorithm for case III. The top left, top right, bottom left and bottom right sub-figures correspond to the 300th, 340th, 420th and 460th frames, respectively.} \label{case3_track}
\end{figure*}

The aforementioned factors together constitute a big challenge for tracking the object accurately online, while the proposed algorithm again gives a satisfactory performance, since it tracks the object very accurately from beginning to end. See Fig.\ref{case3_track} for the tracking results corresponding to the 300th, 340th, 420th and 460th frames of the video.

\begin{figure*}
\begin{tabular}{c}
\centerline{\includegraphics[width=6in,height=3.5in]{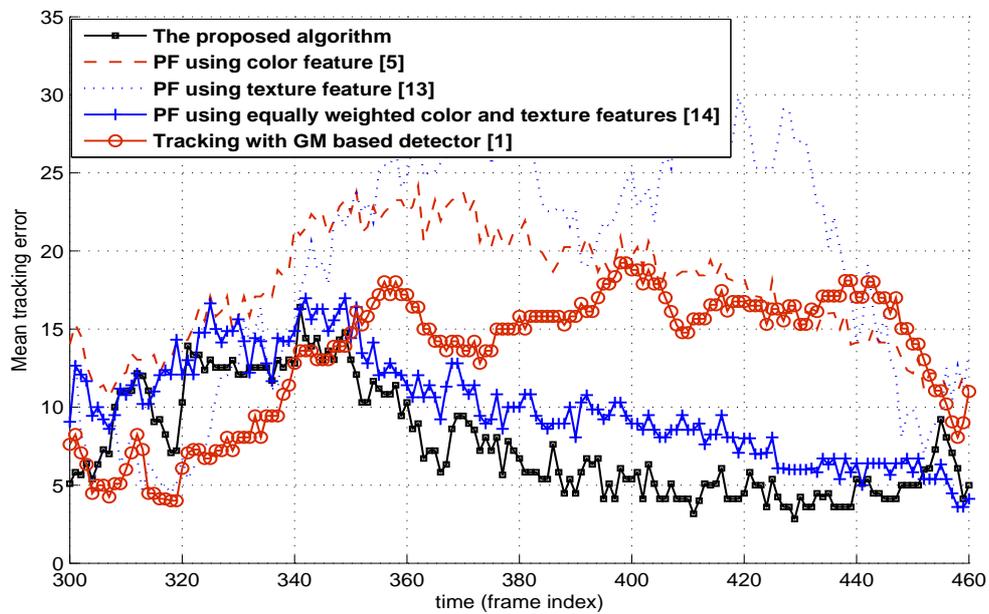}}
\end{tabular}
\caption{Mean tracking error in test case III.} \label{error_case3}
\end{figure*}

\begin{figure*}
\begin{tabular}{c}
\centerline{\includegraphics[width=6in,height=3.5in]{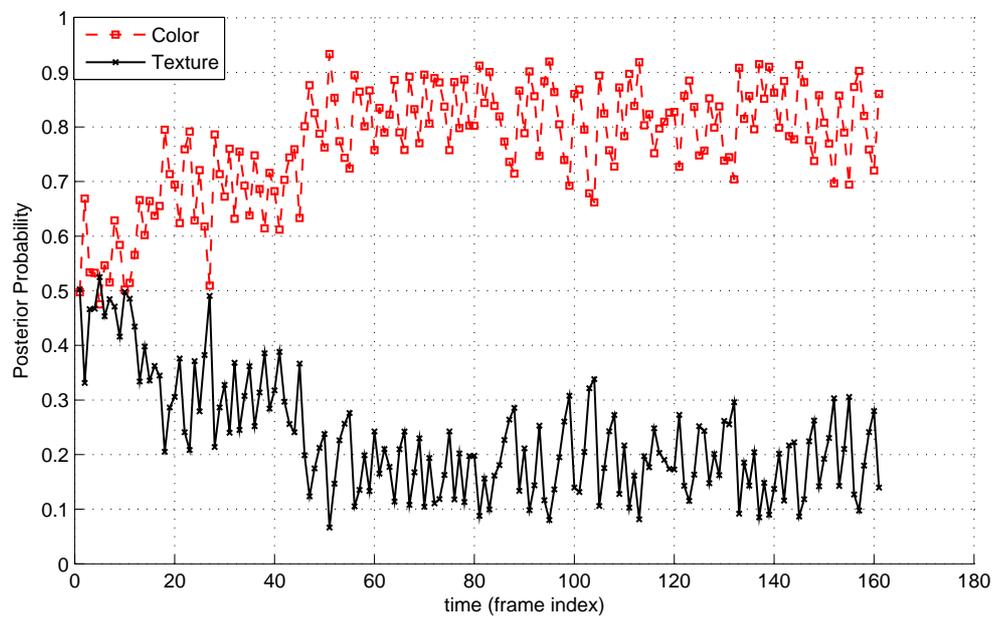}}
\end{tabular}
\caption{Posterior probabilities of the involved feature models at each time step in test case III.} \label{prob_feature_caseIII}
\end{figure*}

A Monte Carlo based numerical performance comparison was also conducted. Every algorithm under consideration was ran 100 times. The result is presented in Fig. \ref{error_case3}. As is shown, the adaptive GM detector \cite{stauffer1999adaptive} performs best in the beginning 40 frames, while since the 340th frame when the route of the object begins to turn, the proposed algorithm gives the best performance in tracking accuracy in most of the time. Fig. \ref{prob_feature_caseIII} presents the posterior probabilities of the color and texture feature models, provided by a typical run of the proposed algorithm. The computation time required to run each algorithm is presented in Table \ref{table_comp2}.

%
%
%
%
%
\section{Conclusion}
In this paper, we propose a BMA based feature fusion approach, along with a generic implementation based on the PF methods, for tracking a moving object of interest in a video stream.
BMA is a theory in Bayesian statistics for dealing with model uncertainty problems. Here we use it to fuse multi-clue information of the object in dealing with complex visual tracking tasks. In theory, the BMA framework allows any number of features to be involved, while as an instantiation, an algorithm that only fuses the color and LBP based texture features is implemented here. We test the performance of the proposed algorithm with real datasets, including the P5 dataset used by PETS 2015 challenge. Our algorithm is shown to be robust against partial occlusion, presence of confusing colors, abrupt changes in the object features and changes in scale. The experimental results show that our algorithm beats several existing competitor algorithms in tracking accuracy with comparable computing burdens. In summarize, we demonstrate that the BMA theory can provide an efficient as well as theoretically sound solution to fuse multi-clue information in visual object tracking.

In this paper, we only use the color and LBP based texture features. Many other possible combinations of differing features can be investigated and used within the BMA framework. It is also feasible to extend the reported method here to handle multi-object visual tracking.
\section{Acknowledgement}
This work was partly supported by the National Natural Science Foundation (NSF) of China under grant Nos. 61302158 and 61571238, the NSF of Jiangsu province under grant No. BK20130869 and the China postdoctoral Science Foundation under grant No. 2015M580455.
\bibliographystyle{IEEEtran}
\bibliography{mybibfile}
\newpage
\begin{table}[tp]
\caption{Computation time comparison in case I(unit: second)}\label{table_comp} \centering \small
\begin{tabular}{c|c|c|c|c}\centering\small\label{table_compare}
       ALG I &ALG II &ALG III &ALG IV &the proposed\\
\hline 3.404&5.668&7.797&9.255&8.632\\
\hline
\end{tabular}
\end{table}
\begin{table}[tp]
\caption{Computation time comparison in case III (unit: second)}\label{table_comp2} \centering \small
\begin{tabular}{c|c|c|c|c}\centering\small\label{table_compare}
       ALG I &ALG II &ALG III &ALG IV &the proposed\\
\hline 12.793&10.286&13.934&22.116&14.637\\
\hline
\end{tabular}
\end{table}
\newpage


\end{spacing}
\end{document}